\documentclass[letterpaper]{article} % DO NOT CHANGE THIS
\usepackage{aaai2027}  % DO NOT CHANGE THIS
\usepackage[hyphens]{url}  % DO NOT CHANGE THIS
\usepackage{graphicx} % DO NOT CHANGE THIS
\usepackage{natbib}  % DO NOT CHANGE THIS AND DO NOT ADD ANY OPTIONS TO IT
\usepackage{caption} % DO NOT CHANGE THIS AND DO NOT ADD ANY OPTIONS TO IT
\usepackage{algorithm}
\usepackage{algorithmic}
\usepackage{amsmath}
\usepackage{amssymb}
\usepackage{bm}
\usepackage{booktabs}
\usepackage{array}
\usepackage{multirow}
\usepackage{newfloat}
\usepackage{listings}
\DeclareCaptionStyle{ruled}{labelfont=normalfont,labelsep=colon,strut=off} % DO NOT CHANGE THIS
\floatstyle{ruled}
\newfloat{listing}{tb}{lst}{}
\floatname{listing}{Listing}

\title{Learning Adaptive Semantic Gaussian Allocation for 3D Occupancy}
\author{
    Kanglin Ning,
    Yiran Zhao,
    Wenrui Li,
    Houde Quan,
    Qifan Li,
    Xingtao Wang,
    Xiaopeng Fan\corresponding
}
\affiliations{
    Harbin Institute of Technology\\
    The Suzhou Research Institute of HIT\\
    The PengChengLab
}

\begin{document}

\maketitle

\begin{abstract}
Semantic 3D Gaussians provide a compact representation for 3D semantic occupancy prediction by rendering semantic primitives into a voxel volume under voxel-wise supervision. Recent methods have improved the modeling ability and efficiency of this representation through more flexible primitive shapes, geometry-guided initialization, and progressive densification. However, these advances mainly determine how primitives are represented, initialized, or added, and do not explicitly address how to select the most useful Gaussians when their total number must be limited to control memory and computation. This imbalance creates an allocation bottleneck: redundant Gaussians remain in simple regions, while difficult regions receive insufficient semantic support. We propose the Semantic Gaussian Allocation Transformer (SAGFormer), which uses Gaussian attributes and local geometric-semantic features to score candidates and select a fixed final Gaussian set. Experiments on nuScenes-SurroundOcc and SSCBench-KITTI-360 show that SAGFormer improves occupancy prediction under the evaluated protocols and yields more semantically consistent and better-utilized Gaussian representations. Under similar final counts and raw coverage, it reduces semantic mixing, strengthens class-consistent voxel support, and produces fewer unused Gaussians. The results indicate that explicit capacity allocation is a useful complement to Gaussian refinement for semantic
occupancy prediction.
\end{abstract}

%     \link{Code}{https://anonymous.4open.science/r/SAGFormer}

\section{Introduction}
3D occupancy prediction estimates the geometry and semantic labels of the surrounding space, including both visible and occluded regions \cite{wei2023surroundocc,tian2023occ3d}. Dense voxel grids provide a direct scene representation, but their memory and computation increase rapidly with spatial range and resolution \cite{li2023voxformer}. Semantic Gaussian methods provide a sparse alternative. They represent a scene with semantic primitives and render these primitives into a voxel occupancy volume \cite{huang2024gaussianformer,zhao2025gaussianformer3d}.

\begin{figure}[t]
    \centering
    \includegraphics[width=1.0\linewidth]{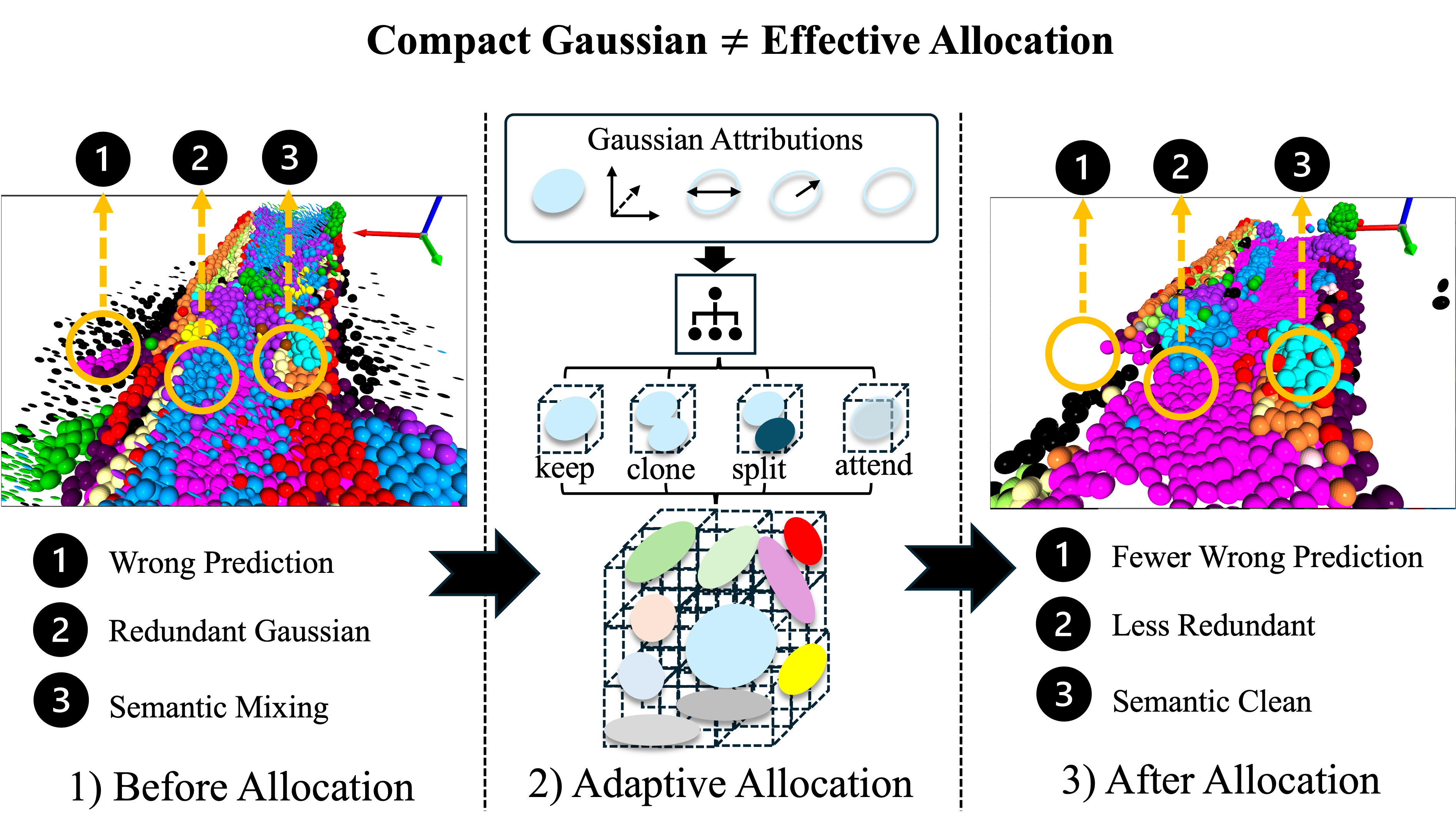}
\caption{
    Semantic Gaussian allocation. Under a fixed Gaussian budget, SAGFormer uses intrinsic and local geometric-semantic cues to select candidates from a shared pool, reducing redundant, semantically mixed, and erroneous Gaussians.
    }
    \label{fig:allocation_positioning}
\end{figure}

Recent work improves sparse primitive representations along three main directions. SplatSSC uses depth cues to construct a compact initial Gaussian set \cite{qian2026splatssc}. QuadricFormer replaces ellipsoidal Gaussians with more flexible superquadrics and applies count-preserving pruning and splitting \cite{zuo2025quadricformer}. PG-Occ progressively adds Gaussian queries to recover missing details for open-vocabulary occupancy \cite{yan2026pgocc}. These methods focus on how primitives are represented, initialized, or progressively added. In contrast, we study which candidates should be kept when the final Gaussian budget is fixed.

A compact Gaussian representation does not necessarily imply an effective allocation. Simple regions may contain several Gaussians with similar spatial support, whereas thin structures and class boundaries may receive insufficient semantically reliable support. Consequently, high geometric coverage can coexist with repeated Gaussians and semantic mixing. We refer to this mismatch as an \emph{allocation bottleneck}. Rendering-oriented allocation is embedded in the photometric Gaussian generator rather than applied to an existing 3D Gaussian set \cite{kim2026f4splat,wan2026splatweaver}. Transferring it to occupancy therefore requires redesigning the allocation unit, semantic heads, and supervision. Semantic occupancy instead requires consistent geometry and class predictions at discrete voxels. Candidate selection for this task should therefore account for semantic uncertainty, local density, geometric overlap, and redundancy.

\begin{table*}[!t]
    \centering
    \caption{
    Comparison with closely related methods. They improve primitive shape, initialization, or progressive densification. SAGFormer studies candidate selection under a fixed final budget.
    }
    \label{tab:closest_work_delta}
    \scriptsize
    \setlength{\tabcolsep}{2.5pt}
    \begin{tabular}{@{}>{\raggedright\arraybackslash}p{2.15cm}>{\raggedright\arraybackslash}p{2.25cm}>{\raggedright\arraybackslash}p{2.55cm}>{\raggedright\arraybackslash}p{2.65cm}>{\raggedright\arraybackslash}p{2.65cm}>{\raggedright\arraybackslash}p{1.85cm}@{}}
        \toprule
        Method & Main goal & When primitives change & Main signal & Count rule & Task \\
        \midrule
        SplatSSC \cite{qian2026splatssc}
        & Compact initial set
        & Initialization
        & Image and depth features
        & Sparse initial set
        & Monocular SSC \\
        QuadricFormer \cite{zuo2025quadricformer}
        & Flexible primitive shape
        & After initial training
        & Superquadric scale
        & Prune and split; count unchanged
        & Closed-set occupancy \\
        PG-Occ \cite{yan2026pgocc}
        & Recover fine details
        & Progressive layers
        & Rendered and reference depth
        & Adds queries at each layer
        & Open-vocabulary occupancy \\
        F4Splat/SplatWeaver \cite{kim2026f4splat,wan2026splatweaver}
        & Generate rendering GS
        & Image-aligned generation
        & Photometric cues
        & Scale threshold / pixel-wise count
        & Novel-view synthesis \\
        SAGFormer (ours)
        & Select useful candidates
        & Post-initialization
        & Geometry, local density, and semantic consistency
        & Per-sample global Top-$K_B$
        & Closed-set occupancy \\
        \bottomrule
    \end{tabular}
\end{table*}

To address this problem, we propose the Semantic Gaussian Allocation Transformer (SAGFormer). It encodes each Gaussian using its attributes and local geometric-semantic context, models relations among Gaussians, and generates keep, clone, split, and suppress candidates. These candidate operations are standard. The main contribution is an occupancy-oriented scoring and selection process that places all candidates in a shared pool. Global Top-$K_B$ selection then keeps a fixed number of candidates for each sample. We further analyze redundancy, semantic mixing, class-consistent voxel support, and Gaussian usage to characterize the resulting allocation.

The main contributions of this work are summarized as follows:

\begin{itemize}
\item We formulate semantic Gaussian allocation as a candidate-selection problem under a fixed final budget. Our analysis shows that high raw coverage can coexist with repeated Gaussians, mixed semantic support, and weak support for occupied voxels.
\item We propose SAGFormer, which combines occupancy-related features, relations among Gaussians, standard candidate operations, and global Top-$K_B$ selection. The hard selection enforces the final Gaussian count for each sample.
\item We evaluate SAGFormer on two semantic occupancy benchmarks. Matched comparisons and allocation measures demonstrate improvements in both occupancy accuracy and the use of the Gaussian budget.
\end{itemize}

\section{Related Work}

\subsection{3D Occupancy Prediction}

Semantic occupancy prediction recovers scene geometry and class labels. MonoScene predicts a complete 3D scene from one image \cite{cao2021monoscene}. Camera-based driving methods use sparse voxel queries, multi-view features, and occupancy supervision from LiDAR or multiple frames \cite{li2023voxformer,wei2023surroundocc,tian2023occ3d}. LiDAR and multi-modal methods use efficient 3D networks, object detection, or semantic and depth cues \cite{roldao2020lmscnet,yang2024daocc,duan2025sdgocc}. These methods are effective. However, the cost of a dense voxel grid grows with the scene range and resolution. This motivates sparse scene representations.

\subsection{Semantic Gaussian-based Occupancy Prediction}

Sparse semantic Gaussians provide a compact alternative to dense voxel grids. GaussianFormer refines a fixed set of Gaussians before voxel decoding \cite{huang2024gaussianformer}. GaussianFormer-2 uses a probabilistic mixture and a data-based initialization to place fewer Gaussians in empty space \cite{huang2024gaussianformer2}. Multi-modal methods combine LiDAR geometry and image semantics through better initialization, attention, feature fusion, or geometry completion \cite{zhao2025gaussianformer3d,doruk2026gaussianocc3d,lv2026gauocc}. Other methods improve relations among Gaussians, temporal information, visual and geometric alignment, collaborative prediction, or dual-Gaussian models \cite{song2025graphgsocc,yan2025stgs,yan2026vg3s,zhou2026generalizing,shi2026odg,chen2026vision}.

Recent methods improve semantic Gaussian representations through depth-guided sparse initialization in SplatSSC \cite{qian2026splatssc}, count-preserving superquadric pruning and splitting in QuadricFormer \cite{zuo2025quadricformer}, or progressive query addition in regions with missing depth evidence in PG-Occ \cite{yan2026pgocc}. SAGFormer does not introduce a new primitive shape or a new depth initializer. It also does not add a fixed group of queries at each layer. It creates several candidates from each input Gaussian and lets all candidates compete for one fixed final budget. The score uses semantic uncertainty and local geometric-semantic context.

\subsection{Gaussian Allocation and Candidate Selection}

Learned Gaussian allocation has mainly been studied for novel-view synthesis. Generative Densification upsamples coarse Gaussian features, whereas F4Splat allocates a global budget from spatial complexity and view overlap and SplatWeaver predicts local counts \cite{nam2024generative,kim2026f4splat,wan2026splatweaver}. These methods embed allocation in image-aligned photometric generation, while others adjust Gaussian placement using image gradients, detections, or reconstruction features \cite{zhang2024pixelgs,ye2024absgs,jiang2025anysplat,moreau2025offthegrid}.

Semantic 3DGS methods use semantic and shape cues for pruning and densification \cite{he2025joint}, semantic gradients for boundary splitting \cite{zhang2025cobgs}, or selective prediction to remove repeated Gaussians \cite{sheng2025spatialsplat}; they target rendering, segmentation, or compact reconstruction. Unlike motion-specific mixture-of-experts routing \cite{jin2026designmoe}, SAGFormer scores candidates derived from initialized semantic primitives using voxel-occupancy evidence and globally selects the final Top-$K_B$ set before decoding. Table~\ref{tab:closest_work_delta} summarizes the differences.

\section{Method}
\label{sec:method}

\subsection{Problem and Overview}
\label{sec:problem}
\begin{figure*}[!t]
\centering
\includegraphics[width=\textwidth]{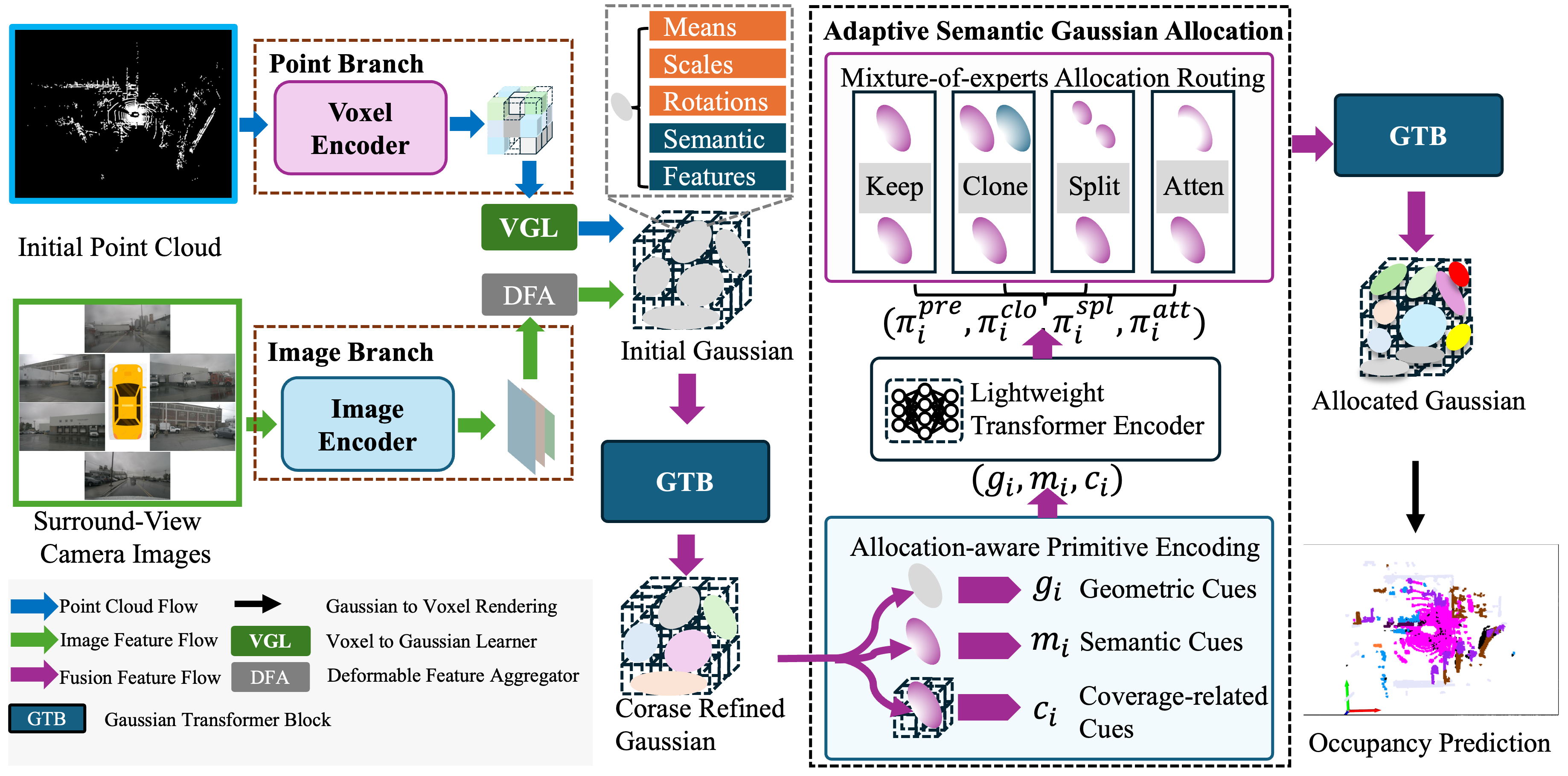}
\caption{
Overall framework of SAGFormer. Multi-modal features initialize semantic Gaussians. The allocation module uses intrinsic attributes and local geometric-semantic context to generate candidates and select the final set before voxel decoding.
}
\label{fig:framework}
\end{figure*}

Given multi-modal inputs, the model starts with initialized Gaussian set $\mathcal{G}^{0}=\{\mathcal{G}^{0}_i\}_{i=1}^{N}$. Each Gaussian is $\mathcal{G}_i^0=\{\bm{\mu}_i,\bm{s}_i,\bm{r}_i,\alpha_i,\bm{f}_i,\bm{p}_i\}$, where the entries denote its center, scale vector, rotation, opacity, feature, and semantic logits, respectively. To address the allocation bottleneck, SAGFormer expands this initial set into a shared candidate pool and ranks the candidates under a normalized budget $B$, with the goal of reducing redundant allocation and strengthening semantically reliable voxel support. The selected set $\hat{\mathcal{G}}$ is then used for voxel decoding.

The budget $B$ controls the allowable growth relative to the initial size $N$. Training uses a differentiable soft expected count, whereas inference materializes a fixed-size set through hard Top-$K$ selection. We define the exact selection rule and its soft training counterpart below. ``Final \#G'' always refers to the number of Gaussians after hard selection, not the soft count used for training.

Figure~\ref{fig:framework} shows the full model. Image and point-cloud features first initialize semantic Gaussians. SAGFormer then builds allocation features and models relations among Gaussians. Finally, it generates keep, clone, split, and suppress candidates. These operations retain reliable support, add nearby support, divide a large or mixed region, or reduce a Gaussian's effect. Global selection decides which candidates enter the decoder.

\subsection{Adaptive Semantic Gaussian Allocation}

Given $\mathcal{G}^{0}$, SAGFormer has three steps: feature encoding, candidate generation and scoring, and global Top-$K_B$ selection.

\noindent\textbf{Allocation-aware Gaussian encoding.}
For each primitive $\mathcal{G}^{0}_i$, we combine its geometric feature $\bm{g}_i$, semantic feature $\bm{m}_i$, and local context cues $\bm{c}_i$ into an allocation feature:
\begin{equation}
    \bm{e}_i
    =
    \mathrm{MLP}_{enc}([\bm{g}_i,\bm{m}_i,\bm{c}_i]).
\end{equation}
The geometric feature contains the normalized center, scale, rotation, and opacity, while the semantic feature contains logits, confidence, and entropy. The four-dimensional local context is computed as follows. Let $\mathcal{N}_i$ contain at most the $K_{\mathrm{nbr}}=8$ nearest Gaussians within radius $R=2\,\mathrm{m}$, let $n_i=\max(|\mathcal{N}_i|,1)$, and define $\bm{q}_{\ell}=\mathrm{Softmax}(\bm{p}_{\ell})$ for every $\ell\in\{i\}\cup\mathcal{N}_i$. For compactness, let $\langle x_{ij}\rangle_i=n_i^{-1}\sum_{j\in\mathcal{N}_i}x_{ij}$ denote a neighborhood average and $\delta_{ij}=\|\bm{\mu}_j-\bm{\mu}_i\|_2/R$ the normalized distance. The context is
\begin{equation}
\begin{aligned}
\bm{c}_i
&=[c_i^{\mathrm{cnt}},c_i^{\mathrm{dst}},c_i^{\mathrm{geo}},c_i^{\mathrm{sem}}],\\
[c_i^{\mathrm{cnt}},c_i^{\mathrm{dst}}]
&=\left[\frac{|\mathcal{N}_i|}{K_{\mathrm{nbr}}},\langle\delta_{ij}\rangle_i\right],\\
[c_i^{\mathrm{geo}},c_i^{\mathrm{sem}}]
&=\left[\langle\eta_{ij}\rangle_i,\langle\bm{q}_i^\top\bm{q}_j\rangle_i\right].
\end{aligned}
\end{equation}
where the scale-normalized geometric overlap is
\begin{equation}
\eta_{ij}=\exp\!\left(-\frac{1}{2}\sum_{d=1}^{3}
\frac{(\mu_j^{(d)}-\mu_i^{(d)})^2}
{(s_i^{(d)})^2+(s_j^{(d)})^2+\epsilon}\right).
\end{equation}
The semantic feature is constructed from the predicted logits $\bm{p}_i$. We denote the softmax distribution by $\bm{q}_i$, its maximum class probability by $\bm{\kappa}_i$, and its class-wise entropy contributions by $\bm{h}_i$:
\begin{equation}
\begin{aligned}
    \bm{m}_i&=[\bm{q}_i,\bm{\kappa}_i,\bm{h}_i],\\
    \bm{q}_i&=\mathrm{Softmax}(\bm{p}_i), \\
    \bm{\kappa}_i&=\max_{c}q_{i,c},\\
    \bm{h}_i&=-\bm{q}_i\odot\log(\bm{q}_i+\epsilon).
\end{aligned}
\end{equation}
Here, $\epsilon$ is a small constant for numerical stability. The four entries of $\bm{c}_i$ encode local neighbor count, mean distance, geometric overlap, and semantic consistency. They are stop-gradient statistics computed only from predicted Gaussian states, without voxel supervision or ground-truth labels. For $C$ semantic classes, their computation costs $O(NK_{\mathrm{nbr}}C)$ and is linear in $N$ for fixed $K_{\mathrm{nbr}}$ and $C$. A Transformer encoder then models interactions among the allocation features:
\begin{equation}
    \{\bm{z}_i\}_{i=1}^{N}
    =
    \mathrm{TransEnc}(\{ [\bm{e}_i, f_i]\}_{i=1}^{N}).
\end{equation}

\noindent\textbf{Candidate generation and scoring.}
For each Gaussian $\mathcal{G}^{0}_i$, a small router predicts weights for the keep, clone, split, and suppress generators:
\begin{equation}
    \bm{\pi}_i
    =
    \mathrm{Softmax}(\mathrm{MLP}_{op}(\bm{z}_i))
    =
    [
    \pi_i^{pre},
    \pi_i^{clo},
    \pi_i^{spl},
    \pi_i^{att}
    ] .
\end{equation}
The superscripts $pre$, $clo$, $spl$, and $att$ denote keep, clone, split, and suppress, respectively, matching Fig.~\ref{fig:framework}. Let $\mathcal{P}$ collect all generated candidates, with $e$ indexing the generator and $k$ a candidate from that generator. Ranking and hard selection are jointly defined by
\begin{equation}
\begin{aligned}
    \mathrm{score}(\mathcal{G}_{i,e,k})
    &=\pi_i^e\alpha_{i,e,k},\\
    K_B&=\lfloor(1+B)N\rfloor,\\
    \hat{\mathcal{G}}
    &=\operatorname{TopK}_{K_B}(\mathcal{P};\mathrm{score}).
\end{aligned}
\end{equation}
During training, all candidates contribute to occupancy decoding through continuous gate weights, enabling end-to-end optimization. Inference uses the same gate-weighted opacity score and only converts the weighted pool into a fixed-size set. Candidates compete individually, so Top-$K_B$ may retain one or both outputs of a clone or split generator.

\noindent\textbf{Candidate generators and soft count.}
Each generator takes $\bm{z}_i$ and predicts parameters for its candidates. The keep generator outputs one residual-refined version of $\mathcal{G}^{0}_i$ to preserve stable support. The clone and split generators additionally create child Gaussians. For a generic child, they predict a center offset $\bm{o}_i$, an unconstrained scale parameter $\tilde{\bm{\eta}}_i$, an opacity residual $\Delta\alpha_i$, and residuals $\Delta\bm{f}_i$ and $\Delta\bm{p}_i$. We convert the scale and opacity updates to valid ranges with $\bm{\beta}_i=\mathrm{softplus}(\tilde{\bm{\eta}}_i)+\epsilon$ and $\alpha'_i=\sigma(\mathrm{logit}(\alpha_i)+\Delta\alpha_i)$. The resulting child is
\begin{equation}
\begin{aligned}
    \mathcal{G}'_i
    =\{&\bm{\mu}_i+\bm{o}_i,
    \bm{\beta}_i\odot\bm{s}_i,
    \bm{r}_i,
    \alpha'_i,\\
    &\bm{f}_i+\Delta\bm{f}_i,
    \bm{p}_i+\Delta\bm{p}_i\}.
\end{aligned}
\end{equation}
Thus, a child shifts the center, positively rescales the Gaussian, preserves its rotation, and updates opacity, feature, and semantic logits. Clone outputs $\mathcal{G}^{0}_i$ together with one instance of $\mathcal{G}'_i$; their scores remain separate, allowing either or both to survive selection. Split outputs two independently parameterized instances, $\mathcal{G}'_{i,1}$ and $\mathcal{G}'_{i,2}$, to represent a large or semantically mixed region. Suppress produces one candidate with unchanged center, scale, rotation, feature, and semantic logits, but replaces its opacity by $\gamma_i\alpha_i$, where $\gamma_i=\sigma(\mathrm{MLP}_{att}(\bm{z}_i))\in(0,1)$. It therefore reduces a Gaussian's contribution without moving or relabeling it.

Keep and suppress each produce one candidate, whereas clone and split each produce two. The normalized soft count first expands over all four branches and then simplifies because the routing weights sum to one:
\begin{equation}
\begin{aligned}
    \bar{\mathcal{C}}_{\pi}
    &=\frac{1}{N}\sum_{i=1}^{N}
    \left(
    \pi_i^{pre}+2\pi_i^{clo}
    +2\pi_i^{spl}+\pi_i^{att}
    \right)\\
    &=1+\frac{1}{N}\sum_{i=1}^{N}
    \left(\pi_i^{clo}+\pi_i^{spl}\right)
    \in[1,2].
    \label{eq:budget_cost}
\end{aligned}
\end{equation}
The base term accounts for one candidate per Gaussian, while clone and split contribute the expected extras. We use $1+B$ as an upper soft target during training and penalize only excess counts. At inference, all generated candidates are globally ranked by $\mathrm{score}(\mathcal{G}_{i,e,k})$, and the $K_B$ highest-scoring Gaussians form $\hat{\mathcal{G}}$, enforcing the exact physical budget.

\subsection{Training Loss and Count Limit}

SAGFormer uses four losses for occupancy prediction, semantic confidence, occupied-voxel recovery, and budget control. The occupancy loss combines cross-entropy and Lovasz-Softmax, with $\lambda_{\mathrm{lov}}$ weighting the latter:
\begin{equation}
    \mathcal{L}_{\mathrm{occ}}
    =
    \mathcal{L}_{\mathrm{ce}}
    +
    \lambda_{\mathrm{lov}}\mathcal{L}_{\mathrm{lov}}.
\end{equation}
We use normalized entropy to make each candidate's semantic prediction less uncertain. Let $\hat N$ be the number of candidates in the weighted training pool, let $C$ be the number of semantic classes, and let $\bm{\rho}_i=\mathrm{Softmax}(\bm{p}_i)$:
\begin{equation}
    \mathcal{L}_{\mathrm{ent}}
    =
    -\frac{1}{\hat{N}\log C}
    \sum_{i=1}^{\hat{N}}
    \sum_{c=1}^{C}
    \rho_{i,c}\log(\rho_{i,c}+\epsilon).
\end{equation}
This entropy term does not directly optimize the support measures in our analysis; those measures are computed only after inference from the ground-truth labels of covered voxels. To reduce false-empty predictions, let $\mathcal{V}_{\mathrm{occ}}$ be the set of occupied ground-truth voxels and $\hat{p}_{v}^{\mathrm{emp}}$ the predicted empty probability at voxel $v$. We define the recovery loss as
\begin{equation}
    \mathcal{L}_{\mathrm{rec}}
    =
    -\frac{1}{|\mathcal{V}_{\mathrm{occ}}|}
    \sum_{v\in\mathcal{V}_{\mathrm{occ}}}
    \log(1-\hat{p}_{v}^{\mathrm{emp}}+\epsilon).
\end{equation}
This loss acts on predicted empty probability rather than directly optimizing the post-hoc coverage measure. The budget loss penalizes a soft count above the target:
\begin{equation}
    \mathcal{L}_{\mathrm{bud}}
    =
    \max(0,\bar{\mathcal{C}}_{\pi}-(1+B)).
\end{equation}
The overall objective combines the four terms, with $\lambda_{\mathrm{ent}}$, $\lambda_{\mathrm{rec}}$, and $\lambda_{\mathrm{bud}}$ controlling their relative weights:
\begin{equation}
    \mathcal{L}
    =
    \mathcal{L}_{\mathrm{occ}}
    +
    \lambda_{\mathrm{ent}}\mathcal{L}_{\mathrm{ent}}
    +
    \lambda_{\mathrm{rec}}\mathcal{L}_{\mathrm{rec}}
    +
    \lambda_{\mathrm{bud}}\mathcal{L}_{\mathrm{bud}}.
\end{equation}
During training, these losses are computed from the weighted candidate pool and its occupancy prediction. Hard Top-$K_B$ selection is used at inference.

\section{Experiments}
\label{sec:experiments}

\begin{table}[t]
    \centering
    \caption{
    Overall benchmark comparison on nuScenes-SurroundOcc. C/L denote camera/LiDAR, and 1/M denote single-/multi-sweep LiDAR.
    $^{*}$ uses one raw LiDAR sweep at inference with temporally aggregated training supervision; $\dagger$ marks reported results, $\ddagger$ our reproduction, and -- unavailable results.
    Full class-wise results are provided in Supplementary Sec.~B.
    }
    \label{tab:nuscenes_main}
    \scriptsize
    \setlength{\tabcolsep}{2.6pt}
    \resizebox{\linewidth}{!}{
    \begin{tabular}{lcccc}
        \toprule
        Method & Mod. & Frames & IoU $\uparrow$ & mIoU $\uparrow$ \\
        \midrule
        GaussianFormer & C & 1 & 29.83 & 19.10 \\
        GaussianFormer-2 & C & 1 & 31.74 & 20.82 \\
        GaussianFormer$^{\ddagger}$ (16.6K) & C & 1 & 29.60 & 19.10 \\
        + SAGFormer allocation (16.6K) & C & 1 & 33.04 & 23.67 \\
        \midrule
        GaussianFormer3D$^{\dagger}$ & C+L & M (10) & 43.3 & 27.1 \\
        GaussianFormer3D$^{\ddagger}$ & C+L & 1 & 36.7 & 22.4 \\
        GaussianOcc3D$^{\dagger}$ & C+L & M & -- & 28.9 \\
        DAOcc$^{\dagger}$ & C+L & M (10) & 45.0 & 30.5 \\
        DAOcc$^{\ddagger}$ & C+L & 1 & 40.1 & 25.6 \\
        SDG-Fusion$^{\dagger}$ & C+L & M & 45.2 & 31.7 \\
        SDG-Fusion$^{\ddagger}$ & C+L & 1 & 41.1 & 26.4 \\
        Gau-Occ$^{\dagger}$ & C+L & $1^{*}$ & 44.3 & 32.7 \\
        SAGFormer w/o Semantic Cues & C+L & 1 & $39.47{\pm}0.45$ & $27.68{\pm}0.12$ \\
        \textbf{SAGFormer} & C+L & 1 & $41.74{\pm}0.11$ & $28.47{\pm}0.06$ \\
        \textbf{SAGFormer} & C+L & $1^{*}$ & $51.79{\pm}0.12$ & $33.16{\pm}0.04$ \\
        \bottomrule
    \end{tabular}
    }
\end{table}

\begin{table}[t]
    \centering
    \caption{
    Overall benchmark comparison on SSCBench-KITTI-360. C/L denote camera/LiDAR.
    $^{*}$ uses one raw LiDAR sweep at inference with temporally aggregated training supervision; $\dagger$ marks reported results, and -- unavailable results.
    Full class-wise results are provided in Supplementary Sec.~B.
    }
    \label{tab:kitti360_main}
    \scriptsize
    \setlength{\tabcolsep}{2.8pt}
    \resizebox{\linewidth}{!}{
    \begin{tabular}{lcccc}
        \toprule
        Method & Mod. & Frames & IoU $\uparrow$ & mIoU $\uparrow$ \\
        \midrule
        MonoScene & C & 1 & 37.87 & 12.31 \\
        VoxFormer & C & 1 & 38.76 & 11.91 \\
        GaussianFormer & C & 1 & 35.38 & 12.92 \\
        GaussianFormer-2 & C & 1 & 38.37 & 13.90 \\
        LMSCNet & L & 1 & 47.53 & 13.65 \\
        GaussianFormer3D & C+L & 10 & 54.5 & 21.3 \\
        Gau-Occ$^{\dagger}$ & C+L & $1^{*}$ & 58.9 & 25.8 \\
        SAGFormer w/o Semantic Cues & C+L & 1 & $56.44{\pm}0.08$ & $23.76{\pm}0.03$ \\
        \textbf{SAGFormer} & C+L & 1 & $57.40{\pm}0.15$ & $24.20{\pm}0.03$ \\
        \textbf{SAGFormer} & C+L & $1^{*}$ & $60.10{\pm}0.09$ & $26.80{\pm}0.02$ \\
        \bottomrule
    \end{tabular}
    }
\end{table}

\subsection{Experimental Setup}

\noindent\textbf{Datasets.}
We select nuScenes-SurroundOcc \cite{caesar2020nuscenes,wei2023surroundocc} and SSCBench-KITTI-360 \cite{liao2022kitti360,li2024sscbench} as complementary benchmarks for testing SAGFormer across different view configurations and scene distributions. We use the official training/validation splits, spatial ranges, voxel resolutions, and semantic categories from prior occupancy work. nuScenes uses six surround-view images and LiDAR points in the LiDAR coordinate system, whereas KITTI-360 follows the SSCBench protocol with the left-view image and the corresponding LiDAR sweep.

\noindent\textbf{Evaluation Metrics.}
We report occupancy IoU, semantic mIoU, and class-wise IoU. We also measure how well the final Gaussians are allocated. Raw Cov. measures the fraction of occupied voxels covered by a Gaussian. Let $\mathcal{V}_i=\{v_j\in\mathcal{V}_{\mathrm{occ}}\mid d_{ij}<\tau_d\}$ be the occupied GT voxels covered by Gaussian $i$, and let $P_i$ and $\hat c_i$ denote their majority-label fraction and majority class. With $N_{\mathrm{valid}}=|\{i:|\mathcal{V}_i|>0\}|$ and $\mathcal{I}_{\rho}=\{i:|\mathcal{V}_i|>0,P_i\ge\rho\}$, we compute
\[
\begin{aligned}
\mathrm{Unused}
&=\frac{1}{N}\sum_i\mathbb{1}[|\mathcal{V}_i|=0],\\
\mathrm{Mix\mbox{-}G}
&=\frac{1}{N_{\mathrm{valid}}}
  \sum_{i:|\mathcal{V}_i|>0}\mathbb{1}[P_i<\rho],\\
\mathrm{Sem\mbox{-}Sup}
&=\frac{1}{|\mathcal{V}_{\mathrm{occ}}|}
  \sum_{v_j\in\mathcal{V}_{\mathrm{occ}}}
  \mathbb{1}[\exists i\in\mathcal{I}_{\rho}:\,
  v_j\in\mathcal{V}_i,\,\hat c_i=y_j].
\end{aligned}
\]
$P_{\mathrm{valid}}$ averages $P_i$ over valid Gaussians, while $P_{\mathrm{pen.}}=(1-\mathrm{Unused})P_{\mathrm{valid}}$ additionally penalizes unused Gaussians. These GT-support measures are used only for post-hoc evaluation and are never provided to the allocation module. We use $\rho=0.6$ and $\tau_d=3$; Supplementary Secs.~A and C provide the full definitions and threshold tests.

\noindent\textbf{Implementation Details.}
All methods use matched encoders, splits, voxelization, augmentation, and evaluation scripts. SAGFormer is trained for 20 epochs on nuScenes and 30 epochs on KITTI-360, and results are averaged over seeds 42, 37, and 91. Complete architecture, optimization, software, and hardware details are provided in Supplementary Sec.~A.

% =========================================================
% Overall semantic Gaussian allocation quality
% =========================================================
\begin{table*}[t]
    \centering
    \caption{
    Allocation and GT-support measures on nuScenes-SurroundOcc.
    Final \#G is the average number of output Gaussians; for budgeted variants, it is measured after Top-$K_B$ selection. FG, BG, and Macro are unweighted class-purity averages.
    $^{\ddagger}$The generator-matched control uses the same keep, clone, split, and suppress candidate families as SAGFormer, but receives no geometric, semantic, or coverage cues and applies no explicit Top-$K_B$ budget constraint.
    }
    \label{tab:overall_allocation_quality}
    \label{tab:category_wise_purity}
    \scriptsize
    \setlength{\tabcolsep}{3.6pt}
    \resizebox{\textwidth}{!}{
    \begin{tabular}{lccccccccc}
        \toprule
        \multirow{2}{*}{Method}
        & \multicolumn{6}{c}{Allocation diagnostics}
        & \multicolumn{3}{c}{GT-support purity} \\
        \cmidrule(lr){2-7}
        \cmidrule(lr){8-10}
        & mIoU $\uparrow$ & Final \#G & Raw Cov. $\uparrow$
        & Unused $\downarrow$ & Mix-G $\downarrow$ & Sem-Sup $\uparrow$
        & FG $\uparrow$ & BG $\uparrow$ & Macro $\uparrow$ \\
        \midrule
        GaussianFormer
        & 19.10 & 25.6K & \textbf{0.9988}
        & 0.5259 & 0.4133 & 0.5012
        & 0.167 & 0.461 & 0.295 \\
        GaussianFormer3D$^{\dagger}$
        & 27.10 & 25.6K & 0.8967
        & 0.1535 & 0.3169 & 0.6311
        & 0.227 & 0.477 & 0.336 \\
        SAGFormer w/o Allocation
        & 27.00 & 10.38K & 0.9108
        & 0.0867 & 0.3031 & 0.6411
        & 0.207 & 0.484 & 0.328 \\
        Generator-matched control$^{\ddagger}$
        & 27.43 & 16.5K & 0.9104 & 0.0854 & 0.2963 & 0.6412
        & 0.199 & \textbf{0.488} & 0.308 \\
        SAGFormer
        & \textbf{28.47} & 16.6K & 0.9103
        & \textbf{0.0785} & \textbf{0.2752} & \textbf{0.6627}
        & \textbf{0.234} & \textbf{0.488} & \textbf{0.345} \\
        \bottomrule
    \end{tabular}
    }
\end{table*}

\noindent\textbf{SAGFormer Configuration.}
SAGFormer uses two Transformer blocks and the default budget $B=0.6$. At inference, it collects all candidates, ranks them by gate-weighted opacity, and keeps the top $K_B=\lfloor(1+B)N\rfloor$. All occupancy and efficiency results use this final set. Supplementary Sec.~A gives the loss weights and full configuration.

\noindent\textbf{Comparison Protocol.}
Our main comparison uses raw single-frame inference and matched baselines. SAGFormer w/o Semantic Cues uses the same initial set, allocation architecture, candidate generators, training protocol, and limit $K_B$. It only removes semantic logits, confidence, and entropy. The completion-enhanced $1^{*}$ setting follows the documented Gau-Occ training schedule \cite{lv2026gauocc}. We report it as an additional result because its public implementation is unavailable. Allocation measures include only methods whose final Gaussian outputs can be inspected. Supplementary Sec.~A gives the full protocol.

\subsection{Benchmark Comparison under Matched and Complementary Protocols}

Tables~\ref{tab:nuscenes_main} and~\ref{tab:kitti360_main} compare SAGFormer with other methods using matched inputs. On nuScenes with raw single-frame input, SAGFormer obtains $41.74{\pm}0.11\%$ IoU and $28.47{\pm}0.06\%$ mIoU. It improves over the matched learned-allocation baseline without semantic features by $2.27$ IoU and $0.79$ mIoU. The two variants use the same initial Gaussian set, allocation architecture, candidate generators, training protocol, and final limit $K_B$. The gain therefore comes from occupancy-specific semantic features, not from using more Gaussians. In a 16.6K camera-only GaussianFormer control, SAGFormer improves IoU/mIoU from $29.60/19.10\%$ to $33.04/23.67\%$ ($+3.44/+4.57$), supporting transfer beyond the multimodal host (Supplementary Secs.~A--B). Under the raw multimodal protocol, SAGFormer also improves over our reproduced GaussianFormer3D and SDG-Fusion results. On KITTI-360, it reaches $57.40{\pm}0.15\%$ IoU and $24.20{\pm}0.03\%$ mIoU.

In the completion-enhanced $1^{*}$ setting, SAGFormer obtains $51.79{\pm}0.12\%$ IoU and $33.16{\pm}0.04\%$ mIoU on nuScenes. It obtains $60.10{\pm}0.09\%$ IoU and $26.80{\pm}0.02\%$ mIoU on KITTI-360. These are additional results and are separate from the main raw single-frame comparison. Supplementary Sec.~B gives the full class-wise results and the source of each result.

\subsection{Allocation Bottleneck Diagnosis}

\begin{figure}[!t]
    \centering
    \includegraphics[width=\linewidth]{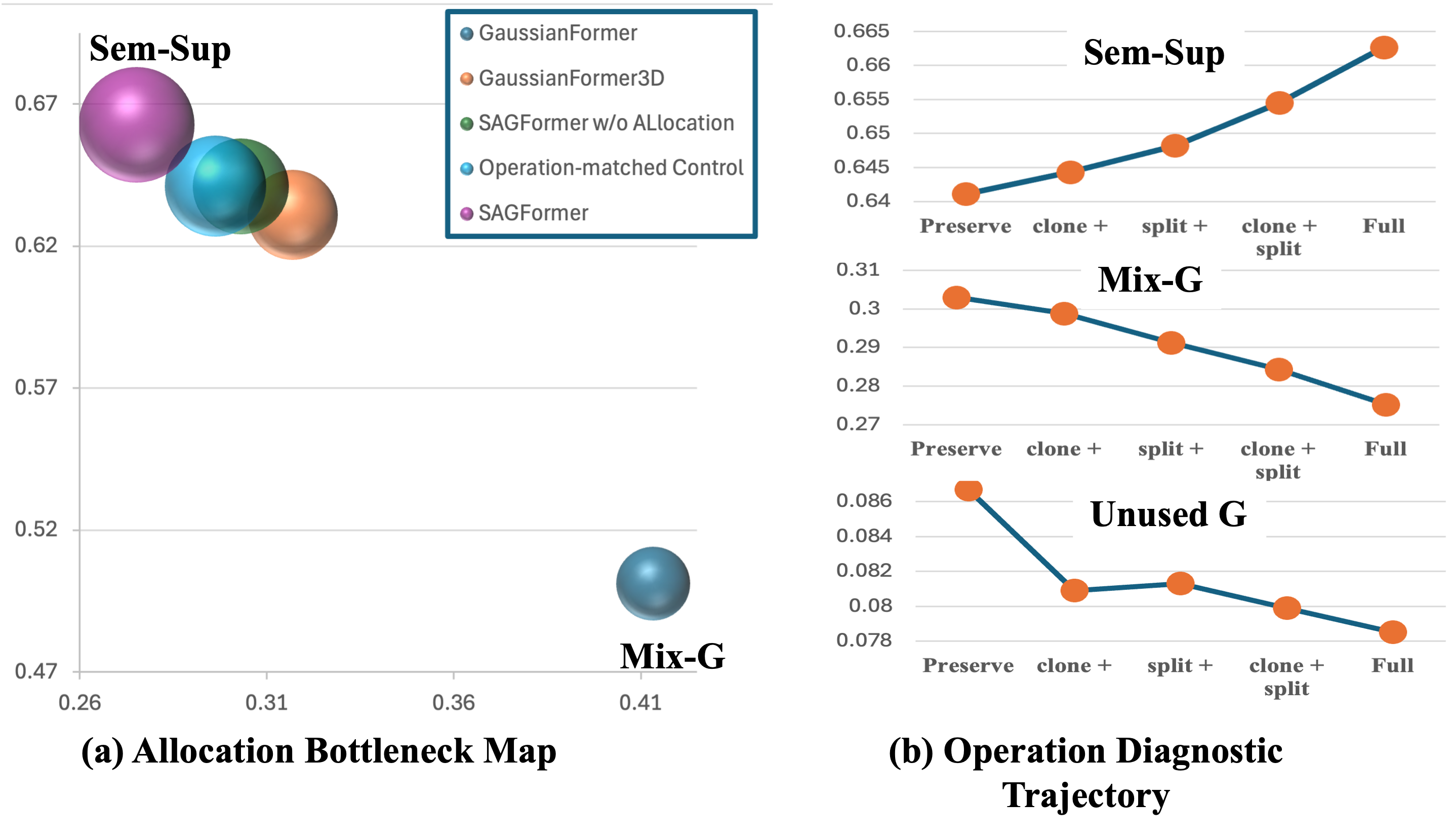}
    \caption{
    Allocation diagnostics on nuScenes-SurroundOcc: (a) method-level Mix-G versus Sem-Sup and (b) candidate-family trajectories for Sem-Sup, Mix-G, and unused ratio.
    Only methods whose final Gaussian outputs can be inspected are included.
    }
    \label{fig:allocation_diagnostics}
\end{figure}

We first test whether the Gaussians are poorly allocated, not only whether occupancy accuracy is low. Table~\ref{tab:overall_allocation_quality} measures three problems under the same voxel-to-Gaussian assignment rule. The unused ratio measures redundant Gaussians. Mix-G measures Gaussians that support mixed classes. Sem-Sup measures class-consistent support for occupied voxels. Figure~\ref{fig:allocation_diagnostics} shows these results. GaussianFormer reaches almost full raw coverage, but $52.59\%$ of its Gaussians are unused. It also has the highest Mix-G ($0.4133$) and lowest Sem-Sup ($0.5012$). Thus, high raw coverage can hide redundant Gaussians and weak semantic support. Compared with the cue-free, budget-unconstrained generator-matched control, SAGFormer has almost the same raw coverage and final count. It reduces Mix-G from $0.2963$ to $0.2752$, increases Sem-Sup from $0.6412$ to $0.6627$, and reduces the unused ratio from $0.0854$ to $0.0785$.

SAGFormer also improves category-balanced GT-support Macro Avg. from $0.328$ to $0.345$ and foreground GT-support purity from $0.207$ to $0.234$. Full class-wise purity, distance-wise quality, and $\tau_d$ sensitivity in Supplementary Sec.~C show consistent gains without relying on extra raw coverage. Overall, the diagnostics connect occupancy gains to reduced redundancy and semantic mixing with stronger class-consistent support.

\FloatBarrier

\subsection{Ablation Studies}

Table~\ref{tab:ablation_allocation_strategy} compares allocation policies with the same initial set, backbone, decoder, training objectives, and 16.6K final budget. Matched QuadricFormer-, F4Splat-, and PG-Occ-style adaptations reach $27.56\%$--$28.03\%$ mIoU. SAGFormer reaches $28.47\%$, exceeding the strongest PG-Occ-style control by $0.44$ mIoU and $0.70$ IoU while raising $P_{\mathrm{pen.}}$ from $0.6553$ to $0.6593$. Random allocation gives only a small gain, and replacing the Transformer with an MLP reduces mIoU to $28.01\%$. These results support gains from the full occupancy-specific allocation policy beyond output capacity and direct transfers of existing allocation rules.

% Required packages:
% \usepackage{booktabs}
% \usepackage{graphicx}
% \usepackage{amssymb}

% =========================================================
% Table 1: Ablation on adaptive allocation strategy
% =========================================================
\begin{table}[!t]
    \centering
    \caption{
    Comparison of matched Gaussian allocation policies on nuScenes-SurroundOcc.
    QuadricFormer-, F4Splat-, and PG-Occ-style policies retain their defining allocation rules but use our initial Gaussian set, occupancy backbone, decoder, training objectives, and input modality; every variant passes exactly 16.6K final Gaussians to the decoder.
    $^{\star}$The capacity-matched w/o Allocation variant here therefore uses a different setting from its counterparts in Tables~\ref{tab:nuscenes_main}, \ref{tab:overall_allocation_quality}, and~\ref{tab:efficiency}.
    }
    \label{tab:ablation_allocation_strategy}
    \scriptsize
    \setlength{\tabcolsep}{2.8pt}
    \resizebox{\linewidth}{!}{
    \begin{tabular}{lcccccc}
        \toprule
        Variant
        & IoU $\uparrow$
        & mIoU $\uparrow$
        & Mean Cov.
        & $P_{\mathrm{valid}} \uparrow$
        & $P_{\mathrm{pen.}} \uparrow$
        & Unused $\downarrow$ \\
        \midrule
        w/o Allocation$^{\star}$
        & $40.4 \pm 0.12$ & $27.33 \pm 0.06$ & 0.9108 & 0.7100 & 0.6484 & 0.0867 \\
        Random Allocation
        & $40.5 \pm 0.13 $& $27.53 \pm 0.05 $ & 0.9102 & 0.7101 & 0.6492 & 0.0833 \\
        QuadricFormer-style \cite{zuo2025quadricformer}
        & $40.5 \pm 0.11$ & $27.56 \pm 0.05$ & 0.9103 & 0.7112 & 0.6497 & 0.0823 \\
        F4Splat-style \cite{kim2026f4splat}
        & $40.9 \pm 0.12$ & $27.89 \pm 0.06$ & 0.9102 & 0.7123 & 0.6503 & 0.0809 \\
        PG-Occ-style \cite{yan2026pgocc} &
        $41.0 \pm 0.12$ & $28.03 \pm 0.05$ & 0.9103 & 0.7134 & 0.6553 & 0.0792 \\
        MLP Controller
        & $41.0 \pm 0.11 $ & $28.01 \pm 0.04$ & 0.9101 & 0.7132 & 0.6544 & 0.0799 \\
        Transformer Controller
        & \textbf{$41.7 \pm 0.11$} & \textbf{$28.47 \pm 0.06$} & 0.9103 & \textbf{0.7155} & \textbf{0.6593} & \textbf{0.0785} \\
        \bottomrule
    \end{tabular}
    }
\end{table}

% =========================================================
% Table 2: Ablation on candidate-generation families
% =========================================================
\begin{table}[!t]
    \centering
    \caption{
    Candidate-family ablation using the nuScenes-SurroundOcc diagnostics in Table~\ref{tab:overall_allocation_quality}.
    }
    \label{tab:ablation_operations}
    \scriptsize
    \setlength{\tabcolsep}{2.2pt}
    \resizebox{\linewidth}{!}{
    \begin{tabular}{lccccccc}
        \toprule
        Variant
        & Clone
        & Split
        & Att.
        & mIoU $\uparrow$
        & Sem-Sup $\uparrow$
        & Mix-G $\downarrow$
        & Unused $\downarrow$ \\
        \midrule
        Preserve only
        & -- & -- & --
        & $27.00 \pm 0.06$ & 0.6411 & 0.3031 & 0.0867 \\
        + Clone
        & $\checkmark$ & -- & --
        & $27.71 \pm 0.04$ & 0.6443 & 0.2988 & 0.0809\\
        + Split
        & -- & $\checkmark$ & --
        & $27.69 \pm 0.04$ & 0.6482 & 0.2913 & 0.0813 \\
        + Clone + Split
        & $\checkmark$ & $\checkmark$ & --
        & $28.05 \pm 0.05$ & 0.6545 & 0.2844 & 0.0799 \\
        Full
        & $\checkmark$ & $\checkmark$ & $\checkmark$
        & \textbf{$28.47 \pm 0.06$} & \textbf{0.6627} & \textbf{0.2752} & \textbf{0.0785} \\
        \bottomrule
    \end{tabular}
    }
\end{table}

% =========================================================
% Table 3: Ablation on metric-guided objectives
% =========================================================
\begin{table}[!t]
    \centering
    \caption{
    Ablation on auxiliary objectives.
    $\mathcal{L}_{\mathrm{ent}}$ is the semantic-entropy regularizer and $\mathcal{L}_{\mathrm{rec}}$ the occupied-voxel recovery loss.
    }
    \label{tab:ablation_objectives}
    \scriptsize
    \setlength{\tabcolsep}{2.6pt}
    \resizebox{\linewidth}{!}{
    \begin{tabular}{lcc|cc}
        \toprule
        Variant
        & $\mathcal{L}_{\mathrm{ent}}$
        & $\mathcal{L}_{\mathrm{rec}}$
        & IoU $\uparrow$
        & mIoU $\uparrow$ \\
        
        \midrule
        Occupancy loss only
        & -- & --
        & $41.1 \pm 0.12$ & $28.12 \pm 0.05$ \\
        + Entropy regularizer
        & $\checkmark$ & --
        & $41.3 \pm 0.11$ & $28.23 \pm 0.06$ \\
        + Occupied-voxel recovery
        & -- & $\checkmark$
        & $41.4 \pm 0.12$ & $28.21 \pm 0.05$ \\
        Full objective
        & $\checkmark$ & $\checkmark$
        & \textbf{$41.7 \pm 0.11$} & \textbf{$28.47 \pm 0.06$}\\
        \bottomrule
    \end{tabular}
    }
\end{table}

% =========================================================
% Merged budget-sensitivity and final-selection table
% =========================================================
\begin{table}[!t]
    \centering
    \caption{
    Budget sensitivity and the gap between soft routing and final selection on nuScenes-SurroundOcc.
    Final \#G is averaged over frames. Soft count ratio is the validation mean of $\bar{\mathcal{C}}_{\pi}$ in Eq.~\eqref{eq:budget_cost}. Selection drop compares soft routing with the final Top-$K_B$ set.
    Results are mean $\pm$ std. where available.
    }
    \label{tab:budget_materialization}
    \footnotesize
    \setlength{\tabcolsep}{2.8pt}
    \resizebox{\linewidth}{!}{
    \begin{tabular}{@{}cccccc@{}}
        \toprule
        \multicolumn{4}{c}{Final-count evaluation}
        & \multicolumn{2}{c}{Final-selection check} \\
        \cmidrule(lr){1-4}
        \cmidrule(lr){5-6}
        $B$
        & \shortstack{Final\\\#G}
        & IoU $\uparrow$
        & mIoU $\uparrow$
        & \shortstack{Soft count\\ratio}
        & \shortstack{Selection\\drop} \\
        \midrule
        0.2
        & 12.46K & 40.90 & 27.69 & 1.23 & 0.321 \\
        0.4
        & 14.53K & $41.20 \pm 0.12$ & $28.24 \pm 0.05$ & 1.41 & 0.132 \\
        0.6
        & 16.6K & $41.74 \pm 0.11$ & \textbf{$28.47 \pm 0.06$} & 1.63 & 0.031 \\
        0.8
        & 18.68K & \textbf{$41.80 \pm 0.12$} & $28.35 \pm 0.06$ & 1.82 & 0.023 \\
        1.0
        & 20.05K & 41.60 & 28.31 & 1.98 & 0.022 \\
        \bottomrule
    \end{tabular}
    }
\end{table}

Table~\ref{tab:ablation_operations} tests each type of candidate. Clone candidates improve Sem-Sup from $0.6411$ to $0.6443$, while split candidates reduce Mix-G to $0.2913$. Using both gives $0.6545$ Sem-Sup and $0.2844$ Mix-G. Adding suppress candidates gives the best mIoU, Sem-Sup, Mix-G, and unused ratio. Thus, the candidate types play different and useful roles.

Table~\ref{tab:ablation_objectives} shows that both auxiliary losses are useful. Using them together gives the best result.Table~\ref{tab:budget_materialization} studies the count limit and the gap between soft routing and final Top-$K_B$ selection. The default $B=0.6$ keeps about $1.6N$ Gaussians and gives the best mIoU. The soft count ratio stays close to $1+B$ for all budgets. The mIoU drop after final selection is $0.031$ at $B=0.6$ and no more than $0.023$ for larger budgets. This small drop shows that the final Top-$K_B$ set behaves much like the soft candidate pool. Supplementary Sec.~D reports purity, coverage, and usage measures. It also shows a small selection gap near the default budget on KITTI-360.

\FloatBarrier

\subsection{Efficiency and Boundary Analysis}

\begin{table}[!t]
    \centering
    \caption{
    Efficiency on nuScenes-SurroundOcc.
    Final \#G is the average number of Gaussians selected after Top-$K_B$; memory is in MB and FPS is measured on the same hardware.
    }
    \label{tab:efficiency}
    \scriptsize
    \setlength{\tabcolsep}{3.0pt}
    \resizebox{\linewidth}{!}{
    \begin{tabular}{lcccccc}
        \toprule
        Method
        & Frames
        & Final \#G
        & Memory
        & FPS
        & IoU $\uparrow$
        & mIoU $\uparrow$ \\
        \midrule
        GaussianFormer
        & 1 & 25.6K & 4733 & 4 & 29.83 & 19.10 \\
        GaussianFormer3D
        & 1 & 25.6K & 6043 & 2 & 36.70 & 22.40 \\
        SAGFormer w/o Allocation
        & 1 & 10.38K & 2189 & 4 & 40.30 & 27.00 \\
        SAGFormer
        & 1 & 16.6K & 2456 & 4 & 41.74 & 28.47 \\
        SAGFormer-M
        & $1^{*}$ & 21.1K & 3542 & 2 & 51.79 & 33.16 \\
        \bottomrule
    \end{tabular}
    }
\end{table}

Table~\ref{tab:efficiency} reports reproducible efficiency comparisons on the same hardware. Adaptive allocation slightly increases memory over the non-allocation variant but improves mIoU from $27.00\%$ to $28.47\%$; compared with GaussianFormer and GaussianFormer3D, SAGFormer uses fewer or comparable primitives with higher accuracy and lower measured memory.

Complete qualitative comparisons are provided in Supplementary Sec.~E.

\FloatBarrier

\section{Limitations and Responsible Use}

SAGFormer is evaluated in calibrated camera-LiDAR settings with supervised occupancy. Relative to the cue-free, budget-unconstrained generator-matched control, macro IoU increases by $1.74$ for static structure ($29.36\%$ to $31.10\%$) but only $0.06$ for dynamic foreground ($26.36\%$ to $26.42\%$), so current evidence does not establish dynamic-object gains under sparse single-frame observations. SAGFormer complements geometric completion and temporal reasoning; deployment requires uncertainty handling and validation under distribution shifts, adverse weather, sensor degradation, and rare traffic scenarios.

\section{Conclusion}

We studied how to allocate a limited number of Gaussians for 3D semantic occupancy prediction. Our analysis finds three main problems: redundant Gaussians, Gaussians that mix several classes, and weak class-consistent support for occupied voxels. SAGFormer uses intrinsic attributes and local geometric-semantic context to decide which Gaussians to keep, clone, split, or suppress. Experiments on nuScenes-SurroundOcc and SSCBench-KITTI-360 show better occupancy prediction and better use of the Gaussian budget. These results suggest that explicit Gaussian allocation is useful alongside Gaussian refinement and geometry completion.

\bibliography{aaai2027}

@inproceedings{caesar2020nuscenes,
  title     = {{nuScenes}: A Multimodal Dataset for Autonomous Driving},
  author    = {Caesar, Holger and Bankiti, Varun and Lang, Alex H. and Vora, Sourabh and Liong, Venice Erin and Xu, Qiang and Krishnan, Anush and Pan, Yu and Baldan, Giancarlo and Beijbom, Oscar},
  booktitle = {Proceedings of the IEEE/CVF Conference on Computer Vision and Pattern Recognition (CVPR)},
  pages     = {11621--11631},
  year      = {2020}
}

@article{liao2022kitti360,
  title   = {{KITTI}-360: A Novel Dataset and Benchmarks for Urban Scene Understanding in 2D and 3D},
  author  = {Liao, Yiyi and Xie, Jun and Geiger, Andreas},
  journal = {IEEE Transactions on Pattern Analysis and Machine Intelligence},
  year    = {2022}
}

@inproceedings{li2024sscbench,
  title     = {{SSCBench}: Monocular 3D Semantic Scene Completion Benchmark in Street Views},
  author    = {Li, Yiming and Li, Sihang and Liu, Xinhao and Gong, Moonjun and Li, Kenan and Chen, Nuo and Wang, Zijun and Li, Zhiheng and Jiang, Tao and Yu, Fisher and Wang, Yue and Zhao, Hang and Yu, Zhiding and Feng, Chen},
  booktitle = {Proceedings of the IEEE/RSJ International Conference on Intelligent Robots and Systems (IROS)},
  year      = {2024}
}

@misc{cao2021monoscene,
  title         = {MonoScene: Monocular 3D Semantic Scene Completion},
  author        = {Cao, Anh-Quan and de Charette, Raoul},
  year          = {2021},
  eprint        = {2112.00726},
  archivePrefix = {arXiv},
  primaryClass  = {cs.CV},
  url           = {https://arxiv.org/abs/2112.00726}
}

@misc{li2023voxformer,
  title         = {VoxFormer: Sparse Voxel Transformer for Camera-based 3D Semantic Scene Completion},
  author        = {Li, Yiming and Yu, Zhiding and Choy, Christopher and Xiao, Chaowei and Alvarez, Jose M. and Fidler, Sanja and Feng, Chen and Anandkumar, Anima},
  year          = {2023},
  eprint        = {2302.12251},
  archivePrefix = {arXiv},
  primaryClass  = {cs.CV},
  url           = {https://arxiv.org/abs/2302.12251}
}

@misc{wei2023surroundocc,
  title         = {SurroundOcc: Multi-Camera 3D Occupancy Prediction for Autonomous Driving},
  author        = {Wei, Yi and Zhao, Linqing and Zheng, Wenzhao and Zhu, Zheng and Zhou, Jie and Lu, Jiwen},
  year          = {2023},
  eprint        = {2303.09551},
  archivePrefix = {arXiv},
  primaryClass  = {cs.CV},
  url           = {https://arxiv.org/abs/2303.09551}
}

@misc{tian2023occ3d,
  title         = {Occ3D: A Large-Scale 3D Occupancy Prediction Benchmark for Autonomous Driving},
  author        = {Tian, Xiaoyu and Jiang, Tao and Yun, Longfei and Mao, Yucheng and Yang, Huitong and Wang, Yue and Wang, Yilun and Zhao, Hang},
  year          = {2023},
  eprint        = {2304.14365},
  archivePrefix = {arXiv},
  primaryClass  = {cs.CV},
  url           = {https://arxiv.org/abs/2304.14365}
}

@misc{huang2024gaussianformer,
  title         = {GaussianFormer: Scene as Gaussians for Vision-Based 3D Semantic Occupancy Prediction},
  author        = {Huang, Yuanhui and Zheng, Wenzhao and Zhang, Yunpeng and Zhou, Jie and Lu, Jiwen},
  year          = {2024},
  eprint        = {2405.17429},
  archivePrefix = {arXiv},
  primaryClass  = {cs.CV},
  url           = {https://arxiv.org/abs/2405.17429}
}

@misc{zhao2025gaussianformer3d,
  title         = {GaussianFormer3D: Multi-Modal Gaussian-based Semantic Occupancy Prediction with 3D Deformable Attention},
  author        = {Zhao, Lingjun and Wei, Sizhe and Hays, James and Gan, Lu},
  year          = {2025},
  eprint        = {2505.10685},
  archivePrefix = {arXiv},
  primaryClass  = {cs.CV},
  url           = {https://arxiv.org/abs/2505.10685}
}

@misc{song2025graphgsocc,
  title         = {GraphGSOcc: Semantic and Geometric Graph Transformer for 3D Gaussian Splating-based Occupancy Prediction},
  author        = {Song, Ke and Wu, Yunhe and Siu, Chunchit and Xiong, Huiyuan},
  year          = {2025},
  eprint        = {2506.14825},
  archivePrefix = {arXiv},
  primaryClass  = {cs.CV},
  url           = {https://arxiv.org/abs/2506.14825}
}

@misc{yan2025stgs,
  title         = {ST-GS: Vision-Based 3D Semantic Occupancy Prediction with Spatial-Temporal Gaussian Splatting},
  author        = {Yan, Xiaoyang and Pei, Muleilan and Shen, Shaojie},
  year          = {2025},
  eprint        = {2509.16552},
  archivePrefix = {arXiv},
  primaryClass  = {cs.CV},
  url           = {https://arxiv.org/abs/2509.16552}
}

@misc{doruk2026gaussianocc3d,
  title         = {GaussianOcc3D: A Gaussian-Based Adaptive Multi-modal 3D Occupancy Prediction},
  author        = {Doruk, Abdullah Enes and Ates, Hasan F.},
  year          = {2026},
  eprint        = {2601.22729},
  archivePrefix = {arXiv},
  primaryClass  = {cs.CV},
  url           = {https://arxiv.org/abs/2601.22729}
}

@misc{lv2026gauocc,
  title         = {Gau-Occ: Geometry-Completed Gaussians for Multi-Modal 3D Occupancy Prediction},
  author        = {Lv, Chengxin and Li, Yihui and Yang, Hongyu and Wang, YunHong},
  year          = {2026},
  eprint        = {2603.22852},
  archivePrefix = {arXiv},
  primaryClass  = {cs.CV},
  url           = {https://arxiv.org/abs/2603.22852}
}

@misc{nam2024generative,
  title         = {Generative Densification: Learning to Densify Gaussians for High-Fidelity Generalizable 3D Reconstruction},
  author        = {Nam, Seungtae and Sun, Xiangyu and Kang, Gyeongjin and Lee, Younggeun and Oh, Seungjun and Park, Eunbyung},
  year          = {2024},
  eprint        = {2412.06234},
  archivePrefix = {arXiv},
  primaryClass  = {cs.CV},
  url           = {https://arxiv.org/abs/2412.06234}
}

@misc{kim2026f4splat,
  title         = {F4Splat: Feed-Forward Predictive Densification for Feed-Forward 3D Gaussian Splatting},
  author        = {Kim, Injae and Kim, Chaehyeon and Bae, Minseong and Joo, Minseok and Kim, Hyunwoo J.},
  year          = {2026},
  eprint        = {2603.21304},
  archivePrefix = {arXiv},
  primaryClass  = {cs.CV},
  url           = {https://arxiv.org/abs/2603.21304}
}

@misc{wan2026splatweaver,
  title         = {SplatWeaver: Learning to Allocate Gaussian Primitives for Generalizable Novel View Synthesis},
  author        = {Wan, Yecong and Li, Fan and Shao, Mingwen and Zuo, Wangmeng},
  year          = {2026},
  eprint        = {2605.07287},
  archivePrefix = {arXiv},
  primaryClass  = {cs.CV},
  url           = {https://arxiv.org/abs/2605.07287}
}

@misc{jiang2025anysplat,
  title         = {AnySplat: Feed-forward 3D Gaussian Splatting from Unconstrained Views},
  author        = {Jiang, Lihan and Mao, Yucheng and Xu, Linning and Lu, Tao and Ren, Kerui and Jin, Yichen and Xu, Xudong and Yu, Mulin and Pang, Jiangmiao and Zhao, Feng and Lin, Dahua and Dai, Bo},
  year          = {2025},
  eprint        = {2505.23716},
  archivePrefix = {arXiv},
  primaryClass  = {cs.CV},
  url           = {https://arxiv.org/abs/2505.23716}
}

@misc{zhang2024pixelgs,
  title         = {Pixel-GS: Density Control with Pixel-aware Gradient for 3D Gaussian Splatting},
  author        = {Zhang, Zheng and Hu, Wenbo and Lao, Yixing and He, Tong and Zhao, Hengshuang},
  year          = {2024},
  eprint        = {2403.15530},
  archivePrefix = {arXiv},
  primaryClass  = {cs.CV},
  url           = {https://arxiv.org/abs/2403.15530}
}

@misc{ye2024absgs,
  title         = {AbsGS: Recovering Fine Details for 3D Gaussian Splatting},
  author        = {Ye, Zongxin and Li, Wenyu and Liu, Sidun and Qiao, Peng and Dou, Yong},
  year          = {2024},
  eprint        = {2404.10484},
  archivePrefix = {arXiv},
  primaryClass  = {cs.CV},
  url           = {https://arxiv.org/abs/2404.10484}
}

@misc{yang2024daocc,
  title         = {DAOcc: 3D Object Detection Assisted Multi-Sensor Fusion for 3D Occupancy Prediction},
  author        = {Yang, Zhen and Dong, Yanpeng and Wang, Jiayu and Wang, Heng and Ma, Lichao and Cui, Zijian and Liu, Qi and Pei, Haoran and Zhang, Kexin and Zhang, Chao},
  year          = {2024},
  eprint        = {2409.19972},
  archivePrefix = {arXiv},
  primaryClass  = {cs.CV},
  doi           = {10.1109/TCSVT.2025.3610634},
  url           = {https://arxiv.org/abs/2409.19972}
}

@misc{duan2025sdgocc,
  title         = {SDGOCC: Semantic and Depth-Guided Bird's-Eye View Transformation for 3D Multimodal Occupancy Prediction},
  author        = {Duan, Zaipeng and Dang, Chenxu and Hu, Xuzhong and An, Pei and Ding, Junfeng and Zhan, Jie and Xu, Yunbiao and Ma, Jie},
  year          = {2025},
  eprint        = {2507.17083},
  archivePrefix = {arXiv},
  primaryClass  = {cs.CV},
  url           = {https://arxiv.org/abs/2507.17083}
}

@inproceedings{roldao2020lmscnet,
  title     = {LMSCNet: Lightweight Multiscale 3D Semantic Completion},
  author    = {Roldao, Luis and de Charette, Raoul and Verroust-Blondet, Anne},
  booktitle = {2020 International Conference on 3D Vision (3DV)},
  year      = {2020},
  doi       = {10.1109/3DV50981.2020.00021},
  url       = {https://doi.org/10.1109/3DV50981.2020.00021}
}

@article{yan2026vg3s,
  title={VG3S: Visual Geometry Grounded Gaussian Splatting for Semantic Occupancy Prediction},
  author={Yan, Xiaoyang and Pei, Muleilan and Shen, Shaojie},
  journal={arXiv preprint arXiv:2603.06210},
  year={2026}
}

@inproceedings{zhou2026generalizing,
  title={Generalizing Visual Geometry Priors to Sparse Gaussian Occupancy Prediction},
  author={Zhou, Changqing and Luo, Yueru and Chen, Changhao},
  booktitle={Proceedings of the IEEE/CVF Conference on Computer Vision and Pattern Recognition},
  pages={28578--28587},
  year={2026}
}

@article{shi2026odg,
  title={Odg: Occupancy prediction using dual gaussians},
  author={Shi, Yunxiao and Zhu, Yinhao and Cai, Herbert and Han, Shizhong and Jeong, Jisoo and Ansari, Amin and Porikli, Fatih},
  journal={Advances in Neural Information Processing Systems},
  volume={38},
  pages={32104--32125},
  year={2026}
}

@inproceedings{chen2026vision,
  title={Vision-Only Gaussian Splatting for Collaborative Semantic Occupancy Prediction},
  author={Chen, Cheng and Huang, Hao and Bagchi, Saurabh},
  booktitle={Proceedings of the AAAI Conference on Artificial Intelligence},
  volume={40},
  pages={2796--2804},
  year={2026}
}

@misc{huang2024gaussianformer2,
  title         = {GaussianFormer-2: Probabilistic Gaussian Superposition for Efficient 3D Occupancy Prediction},
  author        = {Huang, Yuanhui and Thammatadatrakoon, Amonnut and Zheng, Wenzhao and Zhang, Yunpeng and Du, Dalong and Lu, Jiwen},
  year          = {2024},
  eprint        = {2412.04384},
  archivePrefix = {arXiv},
  primaryClass  = {cs.CV},
  url           = {https://arxiv.org/abs/2412.04384}
}

@inproceedings{zuo2025quadricformer,
  title     = {QuadricFormer: Scene as Superquadrics for 3D Semantic Occupancy Prediction},
  author    = {Zuo, Sicheng and Zheng, Wenzhao and Han, Xiaoyong and Yang, Longchao and Pan, Yong and Lu, Jiwen},
  booktitle = {Advances in Neural Information Processing Systems},
  year      = {2025},
  url       = {https://proceedings.neurips.cc/paper_files/paper/2025/hash/4470d0b5795c3721e2a622a249ef2154-Abstract-Conference.html}
}

@inproceedings{yan2026pgocc,
  title     = {Progressive Gaussian Transformer with Anisotropy-aware Sampling for Open Vocabulary Occupancy Prediction},
  author    = {Yan, Chi and Xu, Dan},
  booktitle = {The Fourteenth International Conference on Learning Representations},
  year      = {2026},
  url       = {https://openreview.net/forum?id=mHFaflQv93}
}

@inproceedings{qian2026splatssc,
  title     = {{SplatSSC}: Decoupled Depth-Guided Gaussian Splatting for Semantic Scene Completion},
  author    = {Qian, Rui and Cao, Haozhi and Deng, Tianchen and Yuan, Shenghai and Xie, Lihua},
  booktitle = {Proceedings of the AAAI Conference on Artificial Intelligence},
  volume    = {40},
  pages     = {8520--8528},
  year      = {2026},
  doi       = {10.1609/aaai.v40i10.37803},
  url       = {https://ojs.aaai.org/index.php/AAAI/article/view/37803}
}

@inproceedings{he2025joint,
  title     = {Joint Semantic and Rendering Enhancements in 3D Gaussian Modeling with Anisotropic Local Encoding},
  author    = {He, Jingming and Li, Chongyi and Wang, Shiqi and Kwong, Sam},
  booktitle = {Proceedings of the IEEE/CVF International Conference on Computer Vision (ICCV)},
  pages     = {28354--28363},
  year      = {2025},
  url       = {https://openaccess.thecvf.com/content/ICCV2025/html/He_Joint_Semantic_and_Rendering_Enhancements_in_3D_Gaussian_Modeling_with_ICCV_2025_paper.html}
}

@inproceedings{zhang2025cobgs,
  title     = {{COB-GS}: Clear Object Boundaries in {3DGS} Segmentation Based on Boundary-Adaptive Gaussian Splitting},
  author    = {Zhang, Jiaxin and Jiang, Junjun and Chen, Youyu and Jiang, Kui and Liu, Xianming},
  booktitle = {Proceedings of the IEEE/CVF Conference on Computer Vision and Pattern Recognition (CVPR)},
  pages     = {19335--19344},
  year      = {2025},
  url       = {https://openaccess.thecvf.com/content/CVPR2025/html/Zhang_COB-GS_Clear_Object_Boundaries_in_3DGS_Segmentation_Based_on_Boundary-Adaptive_CVPR_2025_paper.html}
}

@misc{sheng2025spatialsplat,
  title         = {SpatialSplat: Efficient Semantic 3D from Sparse Unposed Images},
  author        = {Sheng, Yu and Deng, Jiajun and Zhang, Xinran and Zhang, Yu and Hua, Bei and Zhang, Yanyong and Ji, Jianmin},
  year          = {2025},
  eprint        = {2505.23044},
  archivePrefix = {arXiv},
  primaryClass  = {cs.CV},
  url           = {https://arxiv.org/abs/2505.23044}
}

@misc{moreau2025offthegrid,
  title         = {Off The Grid: Detection of Primitives for Feed-Forward 3D Gaussian Splatting},
  author        = {Moreau, Arthur and Shaw, Richard and Nazarczuk, Michal and Shin, Jisu and Tanay, Thomas and Zhang, Zhensong and Xu, Songcen and P{\'e}rez-Pellitero, Eduardo},
  year          = {2025},
  eprint        = {2512.15508},
  archivePrefix = {arXiv},
  primaryClass  = {cs.CV},
  url           = {https://arxiv.org/abs/2512.15508}
}

@misc{jin2026designmoe,
  title         = {On the Design of Mixture-of-Experts for Dynamic Gaussian Splatting},
  author        = {Jin, In-Hwan and Mun, Hyeongju and Kim, Joonsoo and Yun, Kugjin and Kong, Kyeongbo},
  year          = {2026},
  eprint        = {2607.08250},
  archivePrefix = {arXiv},
  primaryClass  = {cs.CV},
  url           = {https://arxiv.org/abs/2607.08250}
}


\end{document}